 \documentclass[pmlr,twocolumn,10pt]{jmlr} 

\usepackage{booktabs}
\usepackage{siunitx}

\usepackage[switch]{lineno}

\newcommand{\softAUROC}{\textsc{s-AUROC}}
\newcommand{\softAP}{\textsc{s-AP}}

\usepackage[acronym]{glossaries}
\newacronym{ml}{ML}{machine learning}
\newacronym{ap}{AP}{average precision}
\newacronym{auroc}{AUROC}{area under the receiver operating characteristic curve}
\newacronym{tpr}{TPR}{true positive rate}
\newacronym{fpr}{FPR}{false positive rate}

\newcommand{\equal}[1]{{\hypersetup{linkcolor=black}\thanks{#1}}}

\theorembodyfont{\upshape}
\theoremheaderfont{\scshape}
\theorempostheader{:}
\theoremsep{\newline}

\jmlrworkshop{Machine Learning for Health (ML4H) 2025} 

\title[Clinical Uncertainty Impacts Machine Learning Evaluations]{Clinical Uncertainty Impacts Machine Learning Evaluations}

\author{%
\Name{Simone Lionetti}\equal{Joint first authorship.}$^1$ \Email{simone.lionetti@hslu.ch} \\
\Name{Fabian Gr\"oger}\footnotemark[1]$^{2,1}$ \Email{fabian.groeger@unibas.ch} \\
\Name{Philippe Gottfrois}$^2$ \Email{philippe.gottfrois@unibas.ch} \\
\Name{Alvaro Gonzalez-Jimenez}$^3$ \Email{alvaro.gonzalezjimenez@unibas.ch} \\
\Name{Ludovic Amruthalingam}$^1$ \Email{ludovic.amruthalingam@hslu.ch} \\
\Name{Alexander A.\ Navarini}\equal{Joint last authorship.}$^{2,3}$ \Email{alexander.navarini@usb.ch} \\
\Name{Marc Pouly}\footnotemark[2]$^1$ \Email{marc.pouly@hslu.ch} \\
\addr 
$^1$Department of Computer Science, Lucerne University of Applied Sciences and Arts, Switzerland \\
$^2$Department of Biomedical Engineering, University of Basel, Switzerland \\
$^3$Department of Dermatology, University Hospital of Basel, Switzerland \\
}

\begin{document}

\frenchspacing
\maketitle


\begin{abstract}
Clinical dataset labels are rarely certain as annotators disagree and confidence is not uniform across cases.
Typical aggregation procedures, such as majority voting, obscure this variability.
In simple experiments on medical imaging benchmarks, accounting for the confidence in binary labels significantly impacts model rankings.
We therefore argue that machine-learning evaluations should explicitly account for annotation uncertainty using probabilistic metrics that directly operate on distributions.
These metrics can be applied independently of the annotations' generating process, whether modeled by simple counting, subjective confidence ratings, or probabilistic response models.
They are also computationally lightweight, as closed-form expressions have linear-time implementations once examples are sorted by model score.
We thus urge the community to release raw annotations for datasets and to adopt uncertainty-aware evaluation so that performance estimates may better reflect clinical data.
\end{abstract}

\paragraph*{Data and Code Availability.}
Datasets are publicly accessible and detailed in Section \ref{sec:experiments}.

\paragraph*{Institutional Review Board (IRB).}
IRB approval is not required for our work, as it utilizes existing datasets and annotations without further interaction with human subjects.

\section{Introduction}
\label{sec:intro}


The availability of health-related datasets has grown rapidly \citep{kiryati2021dataset}. 
These resources have been instrumental in advancing \gls*{ml} for healthcare, enabling the community to benchmark methods and accelerate progress \citep{johnson2016mimic,tschandl2018ham10000,irvin2019chexpert}. 
A key barrier to their utility is the uncertainty intrinsic to clinical annotations.
Specifically, even among domain experts, agreement on the presence or absence of a finding is often low, reflecting the ambiguity of medical data and the subjectivity of interpretation \citep{elmore2015diagnostic,krause2018grader}.
To mitigate this, it has become common practice to collect multiple annotations per sample \citep{armato2011lung,irvin2019chexpert,raumanns2021enhance}.


For example in dermatology even histopathology, treated as gold standard diagnosis, achieves only moderate agreement. 
Specifically, in an observational study of 60 melanoma cases across three Spanish hospitals, mean inter-observer agreement gave a Cohen's $\kappa$ around $0.5$ \citep{sanzmotilva2025interobserver}. 
This variability reflects intrinsic ambiguity. 
Forcing deterministic labels obscures it and biases evaluation against uncertainty-aware models.

While multi-annotator designs acknowledge the uncertainty of the labeling process, the resulting annotations are typically aggregated into a single ``ground-truth'' label, often by majority voting or thresholding \citep{snow2008cheap}. 
This produces an illusion of certainty: An image labeled as ``positive'' by $6/10$ experts is treated identically to one unanimously labeled as ``positive'' by 10/10.
Even in the case of 2-5 annotators, which is more common in health where expert annotations are costly, a $2/3$ agreement conveys critically different information than a unanimous $3/3$ agreement.
This nuance is lost with majority voting or thresholding.
Current \gls*{ml} pipelines often evaluate models against aggregated labels as if they were certain, disregarding the underlying uncertainty \citep{irvin2019chexpert,chen2021evaluation}.
In doing so, meaningful variations in expert opinion are collapsed into a binary outcome, obscuring the fact that evaluation is performed against a fragile construct rather than a real reference.

We argue that this practice is not aligned with the nuances of clinical data. 
Evaluation should incorporate uncertainty, as ignoring it leads to a hidden selection bias, where models that more closely align with thresholded labels are favored over those that predict realistic uncertainty.
Importantly, doing so is neither conceptually nor technically difficult, and has a sizeable impact on results.
There exist extensions of widely used metrics, such as \gls*{auroc} or \gls*{ap}, that apply to probabilistic labels. 
These can be traced back at least 20 years in the information retrieval literature \citep{kekalainen2002using}, but remain rarely applied in the \gls*{ml} for health community.

\looseness=-1
Rather than collapsing disagreement to a binary label, uncertainty-aware \emph{soft} metrics directly operate on continuous probabilities in the $[0,1]$ range. 
Two properties make them immediately practical. 
First, soft metrics are agnostic to the assumptions used to obtain probability estimates, which can range from independent votes or subjective confidence to item-response theory.
Second, they are computationally tractable, as closed-form expressions allow for linear-time execution after sorting by score, just as in the binary case.

\textbf{Related work.}
The issue of annotation uncertainty and its impact on \gls*{ml} model evaluation, especially for clinical tasks, has been highlighted several times.
\citet{maier2018rankings} showed that rankings in biomedical image analysis challenges fluctuate with annotator selection and aggregation scheme, urging for transparency with respect to label uncertainty.
For clinical applications, \cite{chen2021evaluation} argued that when the reference standard is subjective, agreement should be measured with human comparators, avoiding claims of accuracy against unquestioned truth.
\citet{gordon2021disagreement} took the program further by addressing intra-annotator variations and then averaging metrics across annotators.
A community perspective \citep{reinke2024understanding} emphasized that metric choice and aggregation must align with the problem and data, highlighting pitfalls that arise when subjectivity is ignored.

\looseness=-1
Several thresholdless metrics have been proposed to address uncertain annotations.
The information retrieval formulation of precision and recall to continuous, non-binary labels included the soft version of \gls*{ap} \citep{kekalainen2002using}.
A similar approach was used to evaluate boundary detection in images by taking the average of scores over different annotators \citep{martin2004learning}.
Further works formulated precision and recall with frequencies as probabilities, including all-negative and all-positive dummies to avoid certainty when labels are classifier outputs, and investigated hypothesis testing on these metrics \citep{lamiroy2011computing,lamiroy2015statistical}.

\looseness=-1
\textbf{Contributions.}
This paper draws attention to the gap between annotation practice and evaluation methodology.
Through evaluations on benchmark medical imaging datasets, we show that accounting for label uncertainty can substantially alter the ranking of competing methods. 
This evidence underscores the need for a shift away from evaluation with artificially certain labels towards faithful representation of annotation uncertainty.
We compile explicit, simple expressions to compute soft versions of \gls*{ap} and \gls*{auroc} that are cheap to compute and straightforward to interpret, thus facilitating their adoption.
We conclude by urging the community to follow uncertainty-aware evaluation practices and to promote transparency by releasing unaggregated annotations.

\begin{figure*}[t]
    \centering
    \begin{minipage}[t]{0.6\textwidth}
        \vspace{0pt}
        \centering
        \includegraphics[width=\linewidth]{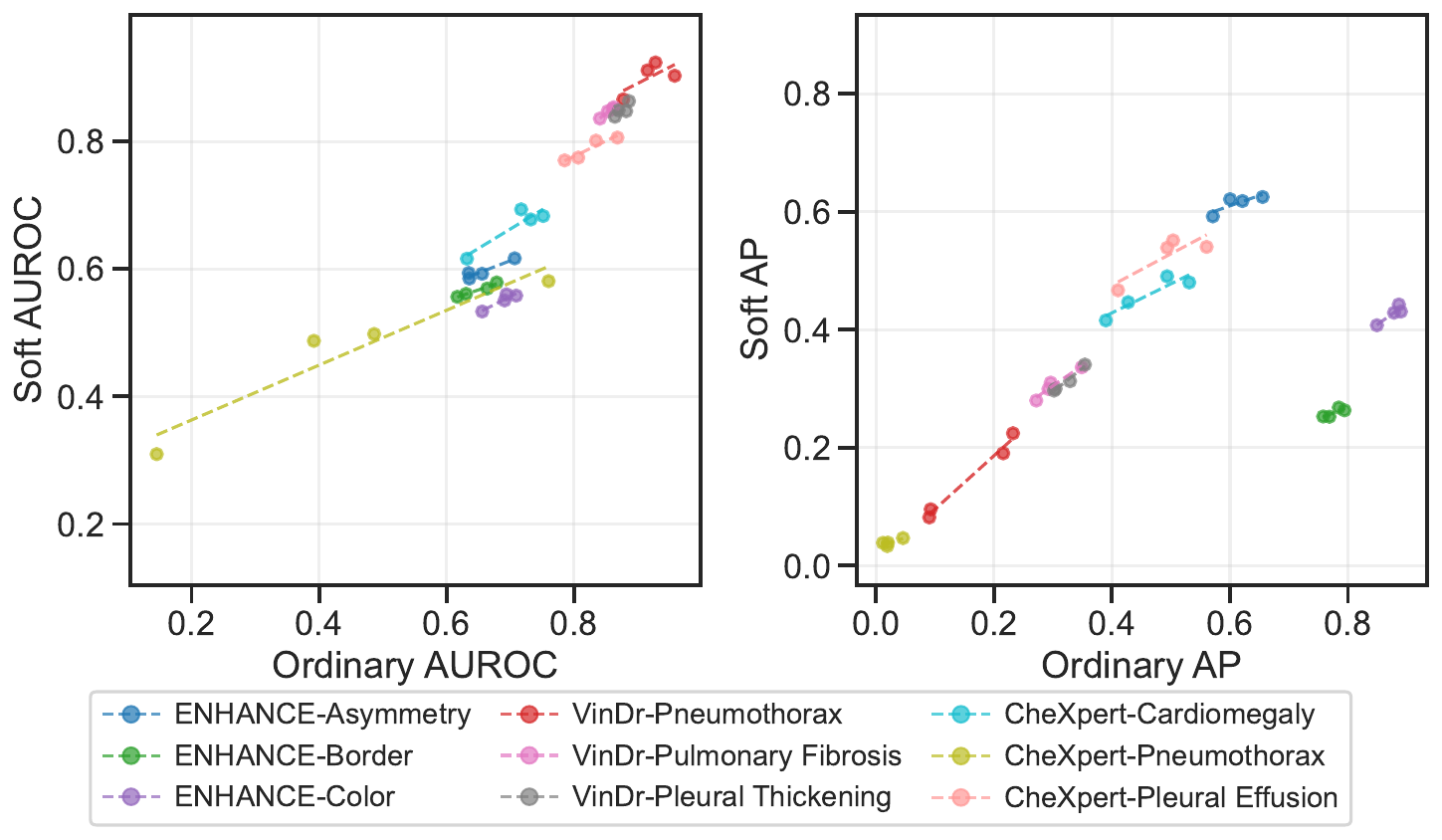}
    \end{minipage}
    \hfill
    \begin{minipage}[t]{0.38\textwidth}
        \vspace{3pt}
        \centering
        \resizebox{\linewidth}{!}{%
        \label{tab:Correlation}
        \begin{tabular}{lcc}
        \toprule
        & \multicolumn{2}{c}{$R^2$} \\
        \cmidrule(lr){2-3}
        \textbf{Dataset} & \textbf{AUROC} & \textbf{AP} \\
        \midrule
        \textit{ENHANCE}-Asymmetry & 0.899 & 0.697 \\
        \textit{ENHANCE}-Border & 0.966 & 0.695 \\
        \textit{ENHANCE}-Color & 0.874 & 0.834 \\
        \textit{VinDr}-Pleural Thickening & 0.697 & 0.969 \\
        \textit{VinDr}-Pneumothorax & 0.465 & 0.983 \\
        \textit{VinDr}-Pulmonary Fibrosis & 0.982 & 0.941 \\
        \textit{CheXpert}-Cardiomegaly & 0.840 & 0.856 \\        
        \textit{CheXpert}-Pleural Effusion & 0.897 & 0.726 \\
        \textit{CheXpert}-Pneumothorax & 0.916 & 0.659 \\
        \bottomrule
        \end{tabular}%
        }
    \end{minipage}
    \caption{
        Comparison of ordinary and soft metrics on three datasets with three tasks each.
        In the left panels, dots represent different backbones, and dashed lines indicate Pearson score correlations whose $R^2$ values are reported on the right.
        Details are in Appendix Table~\ref{tab:OrdinaryVSSoftMetrics}.
    }
    \label{fig:DatasetAnalysis}
\end{figure*}

\section{Experiments}
\label{sec:experiments}

\subsection{Experimental setup}

\textbf{Metrics.}
Let items $1,\dots,n$ be sorted in descending order by their score for positive classification. 
Denote the probability of item $i$ being positive by $p_i\in[0,1]$.
Define the cumulative counts
\begin{equation}
    n_i^+ = \sum_{j=1}^{i} p_j,\qquad
    n_i^- = \sum_{j=1}^{i} (1-p_j),
\end{equation}
and totals $n^+=n_n^+$, $n^-=n_n^-$.
\Gls*{tpr} and \gls*{fpr} are defined by
\begin{equation}
    \mathrm{TPR}_i = n_i^+/n^+,\qquad
    \mathrm{FPR}_i = n_i^-/n^-,
\end{equation}
and precision and recall by
\begin{equation}
    \mathrm{P}_i=n_i^+/i,\qquad
    \mathrm{R}_i=n_i^+/n^+.
\end{equation}

\begin{definition}[Soft \gls*{auroc}]
\label{def:s-auroc}\newline
The soft (uncertainty-aware) \gls*{auroc} is defined as
\begin{equation}
\begin{aligned}
    \mathrm{s\text{-}AUROC}
	&=
    \sum_{i=1}^n \mathrm{TPR}_i(\mathrm{FPR}_i - \mathrm{FPR}_{i-1})
    \\&=
    \sum_{i=1}^n \sum_{j=1}^{i-1} \frac{(1-p_i)p_j}{n^+n^-}.
\end{aligned}
\end{equation}
\end{definition}

\begin{definition}[Soft \gls*{ap}]
\label{def:s-ap}\newline
The soft (uncertainty-aware) \gls*{ap} is defined as
\begin{equation}
    \mathrm{s\text{-}AP}
    =
    \sum_{i=1}^n \mathrm{P}_i(\mathrm{R}_i - \mathrm{R}_{i-1})
    =
    \sum_{i=1}^n \sum_{j=1}^i \frac{p_i}{i}\frac{p_j}{n^+}.
\end{equation}
\end{definition}

These definitions automatically reduce to ordinary \gls*{auroc} and \gls*{ap} when labels are binary.
Using the cumulative sums in the first lines of the equations allows for linear time implementations, assuming the samples are already sorted by score.
The metric values are deterministic and do not require Monte Carlo sampling, despite taking into account the probabilistic nature of the annotation process.

In the experiments that follow, we compare \emph{ordinary} \gls*{auroc} and \gls*{ap} computed against \emph{binary} labels, and their \emph{soft} counterparts \softAUROC\ and \softAP\ computed directly from \emph{probabilistic} labels.

\textbf{Tasks and datasets.}
We first evaluate soft ranking metrics across three medical imaging settings with varying label uncertainty (see Appendix~\ref{apd:uncertainty}).
The first is the dermatology benchmark \textit{ENHANCE} \citep{raumanns2021enhance}, featuring annotations of the student groups 2017--2020 for the lesion attributes asymmetry, border, and color.
For each image, we normalize ratings to $[0, 1]$ and average them to obtain a soft label, and binarize labels for presence of the attribute by requiring a mean $>\!1$ on the original scale. 
We then consider the two chest X-ray benchmarks \textit{VinDR} \citep{nguyen2022vindr} and \textit{CheXpert}\footnote{Only the test set contains multi-rater annotations.} \citep{irvin2019chexpert}.
Here, we take the mean of expert annotations for each image to obtain soft targets and binarize them with majority voting, which aligns with prior work on this dataset~\citep{irvin2019chexpert}.
For datasets without a predefined test split, we select a 30\% test set with a fixed random seed.

To evaluate how uncertainty affects evaluation across a broader range of tasks beyond typical medical imaging, and beyond simple averaging for probabilistic labels, we also investigate data quality issue detection on \textit{CleanPatrick} \citep{groger2025cleanpatrick}.
This benchmark features raw, unaggregated annotations for three issue types on medical images (off-topic samples, near duplicates, and label errors).
We follow the same setup and evaluation as the benchmark, including the evaluated methods and GLAD aggregation (see Appendix~\ref{apd:data-quality-methods}).

\textbf{Features and models.}
We extract image representations with four backbones pretrained on ImageNet with supervision: VGG-16 \citep{simonyan2014very}, ResNet-50 \citep{he2016deep}, EfficientNet-b0 \citep{tan2019efficientnet}, and ViT-base \citep{dosovitskiy2020image}. 
Following the methods' preprocessing, all images are resized to $224\times224$, converted to tensors, and normalized with ImageNet mean and standard deviation. 
\textit{VinDr}'s DICOM files are read using \texttt{pydicom} \citep{pydicom}. 
Pixel arrays are scaled to 8-bit if needed, and single-channel images are repeated to RGB. 
We train a standard Logistic Regression classifier from scikit-learn on the features extracted with frozen backbones and report probabilities on the held-out test set. 
While more sophisticated and performant backbones exist, these models remain relevant in practice for clinical settings, where computational infrastructure may be limited.

\begin{figure}[t]
    \centering
    \includegraphics[width=1.0\linewidth]{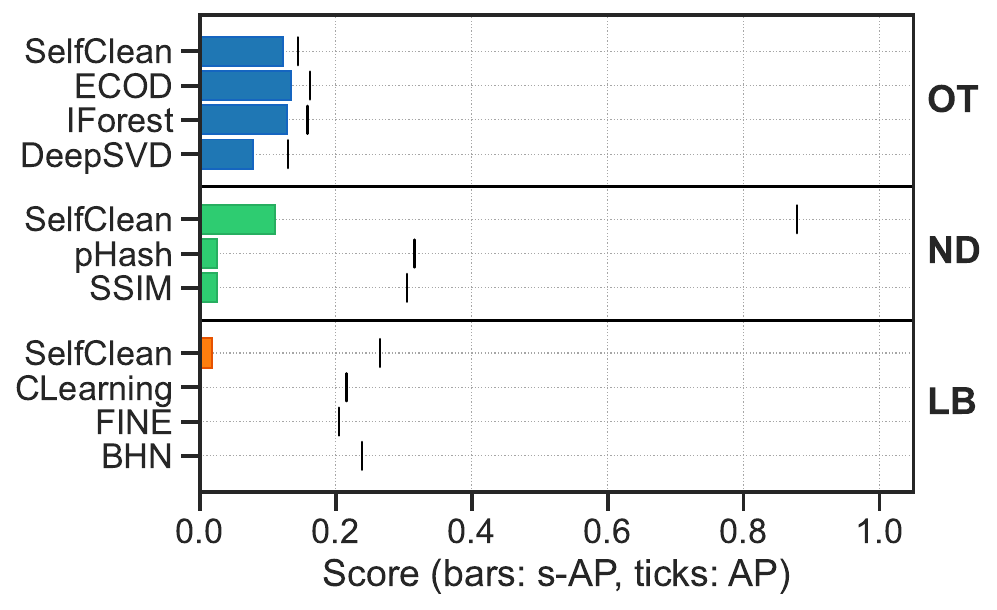}
    \caption{
        Results for data quality issue detection on \textit{CleanPatrick}.
        Bars show the uncertainty-aware score s-AP, and vertical ticks mark the corresponding AP. 
        Detailed results are in Table \ref{tab:OrdinaryVSSoftMetrics} of the appendix.
    }
    \label{fig:cleanpatrick}
\end{figure}

\subsection{Results}

Switching from ordinary to uncertainty-aware evaluation reweights ambiguous cases.
Besides decreasing absolute scores and compressing their range, this importantly reshuffles leaders across datasets as can be seen in Figure~\ref{fig:DatasetAnalysis}. 
On \textit{ENHANCE} Border/Color, ordinary \glspl*{ap} above 0.75 collapse to lower and tighter s-AP regions. 
In this regime, VGG-16 outperforms other backbones, indicating that the best method does not merely separate binary positives, but rather correctly handles borderline lesions.
A similar pattern can be observed on \textit{VinDr}, where soft metrics are uniformly lower yet still flip winners on several tasks (\textit{e.g.}, VGG-16 performs best on Pneumothorax, Pulmonary Fibrosis, and Pleural Thickening), underscoring that rank is sensitive to how label uncertainty is modeled. 
The analysis of the coefficients of determination $R^2$ in Figure~\ref{fig:DatasetAnalysis} shows that \gls*{ap} often decouples rankings between ordinary and soft settings moderately, and \gls*{auroc} can sometimes decouple them sharply (\textit{e.g.}, CheXpert--Pneumothorax AUROC $R^{2}\!=\!0.465$ vs.\ AP $R^{2}\!=\!0.983$).

Note that most top-ranked solutions in the \textit{VinDr} challenge ensemble multiple models, often including pretrained and fine-tuned ResNets. 
Even with frozen backbones and a linear head, VGG-16 achieves better results than ResNet counterparts when evaluated with uncertainty-aware metrics on several tasks.
This suggests that its representations may lead to better logit calibration for evaluations that explicitly model annotation uncertainty, compared to the widely used fine-tuned ResNet variants.

\begin{figure}[t]
    \centering
    \includegraphics[width=1.0\linewidth]{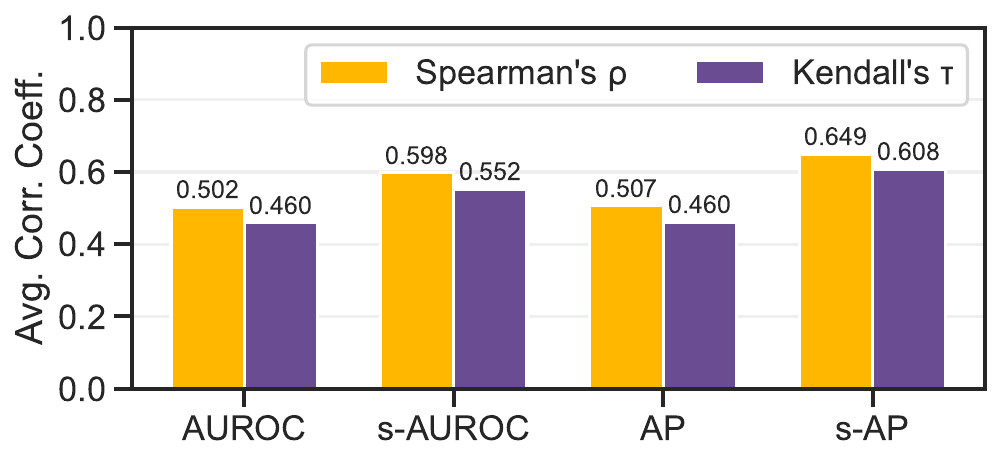}
    \caption{
        Average correlation coefficient of rankings produced by ordinary and soft metrics upon annotation bootstrap across 13 tasks.
        Details are in Figure \ref{fig:bootstrap} of the appendix.
    }
    \label{fig:stability-avg}
\end{figure}

On \textit{CleanPatrick} (see Figure~\ref{fig:cleanpatrick}), uncertainty-aware evaluation does not change the best method but reorders competing ones across tasks.
For \textit{near duplicates}, SelfClean remains first, while pHash and SSIM collapse to a tie under s-AP. 
For \textit{off-topic samples}, the ordering is mostly unchanged. 
The starkest shift is observed for \textit{label errors}: BHN sits second in terms of ordinary \gls*{ap} but drops near the bottom under s-AP, while SelfClean keeps first rank. 
Overall, shifting from ordinary to soft metrics changes which models are better than others in ambiguous cases.

In Figure \ref{fig:stability-avg}, we further investigate the reliability of the ordinary and soft metrics. 
We perform a bootstrap analysis as discussed in Appendix~\ref{apd:bootstrap-analysis} and compare the stability of the model rankings by measuring the correlation coefficient among them.
The results show that using soft metrics leads to more consistent evaluation across all tasks.

To ensure that findings generalize beyond linear evaluation on frozen features, we also conduct experiments on Chexpert both with end-to-end fine-tuning (Table~\ref{tab:finetune-results}) of the same model setup (i.e., backbone and linear head) and using leaderboard models (Table~\ref{tab:chexternal-reranking}). 
These results, detailed in Appendix~\ref{apd:ext-results}, confirm that uncertainty-aware metrics frequently reshuffle model rankings including the top position.

\section{Discussion}
\label{sec:discussion}
This work considers uncertainty in ranking metrics for binary classification. 
While this also applies to binary multi-label problems, extension to the multi-class case is only straightforward in a one-vs-one or a one-vs-all fashion.

Soft metrics are agnostic both to the modality and to the origin of probabilistic labels. 
They are directly applicable to any domain whenever probabilistic labels are available. 
Beyond the example of melanoma diagnosis \citep{elmore2017pathologists}, other clinical domains show high inter-rater variability where soft metrics are advantageous.
In radiology, the interpretation of ambiguous lung nodules or characterization of lesions in mammography often results in significant disagreement \citep{lehman2015diagnostic}. 
In pathology, the grading of tumors, like Gleason scoring for prostate cancer, is notoriously subjective \citep{egevad2013standardization}. 
In ophthalmology, severity grading of diabetic retinopathy leads to expert disagreement \citep{krause2018grader}. 
These can be contrasted with domains where disagreement is typically lower, such as bone fracture detection \citep{nowroozi2024artificial}, to identify where uncertainty-aware evaluation is most critical.

The choice of aggregation model (e.g., simple averaging or item-response theory) may have a significant influence. 
These nuances are not fully explored yet, due to scoping and the limited availability of public medical datasets with unaggregated annotations. 
The work by \cite{stutz2023conformal} offers a complementary perspective to soft ranking metrics, as it describes how to produce prediction sets with confidence guarantees using conformal prediction. 

One might also hypothesize that uncertainty-aware training will produce better results in terms of soft metrics, but evidence for this is yet to be collected. 
The surprising performance of VGG-16 in some tasks warrants discussion. 
One hypothesis is that this architecture may be less prone to overfitting binarized majority labels in mid-small datasets compared to more complex ones. 
Although monotonic calibration such as Platt scaling does not change the result of ranking metrics, models that are not overly confident and incorporate better uncertainty estimates are rewarded by disagreement-aware evaluation.

The full clinical impact of soft-metrics remains to be assessed. 
Results suggest that decisions about which models should be deployed are likely to change due to rank reshuffling. 
While incorporating uncertainty within evaluations is intuitive, downstream consequences should be investigated. 

Finally, releasing unaggregated annotations may in some cases present challenges related to privacy, ethics, or legality, as discussed in Appendix~\ref{apd:barriers}. 
On the other hand, it has great potential to improve fairness, for instance, by investigating if high-disagreement cohorts correlate with demographic factors.

\section{Conclusion}
\label{sec:conclusion}

Label uncertainty is intrinsic to clinical data. 
Uncertainty-aware ranking metrics that operate on probabilistic targets are easy to compute, improve the stability of rankings, and often change which models perform best. 

\looseness=-1
Two practical recommendations follow. 
First, benchmark creators should release unaggregated annotations or at least fractional targets to enable uncertainty-aware evaluation. 
Second, practitioners should report uncertainty-aware metrics alongside ordinary ones and comment on any rank changes. 
These steps make empirical claims more robust to the irreducible ambiguity of clinical annotation.

\section*{Acknowledgments}

This research was funded in part by the Swiss National Science Foundation (SNSF) under grant 20HW-1\_228541.

\bibliography{jmlr-sample}

\appendix

\section{Barriers to data sharing}\label{apd:barriers}
In many practical cases, releasing individual annotations with anonymized annotator identifiers entails low ethical, legal, or privacy risks. 
Annotators are usually professionals participating with informed consent rather than vulnerable subjects, and the additional metadata does not expose new health information beyond what is already available. 
When identifiers are randomized and links to institutions or timestamps are loose, re-identification of either patients or annotators requires substantial prior knowledge and effort. 
Certain datasets may warrant additional precautions due to specific contextual factors, but this is not the norm. 
Provided the release agreement forbids misuse and a standard data-protection assessment confirms anonymization, the potential for harm remains remote, whereas the gains for clinical research are substantial.

\section{Uncertainty distributions}\label{apd:uncertainty}
Figure~\ref{fig:uncertainty-data} shows the distributions of soft and hard labels for the evaluated datasets of the main paper.
We can see that \textit{ENHANCE} has the highest uncertainty, which can be attributed to the large amounts of annotations per sample.
In this case, collapsing the labels to a binary outcome results in the loss of a significant amount of information.
\textit{CheXpert} contains eight annotations per sample, while for \textit{VinDr} we found three annotations per sample, resulting in less uncertainty.

\begin{figure*}[htbp]
\floatconts
  {fig:uncertainty-data}
  {\caption{
    Distribution of soft labels for the evaluated datasets.
    Dotted lines represent the threshold used to produce the hard labels, and the colors represent binary positive and negative labels.
  }}
  {%
    \subfigure[ENHANCE-Asymmetry]{%
      \includegraphics[width=0.32\linewidth]{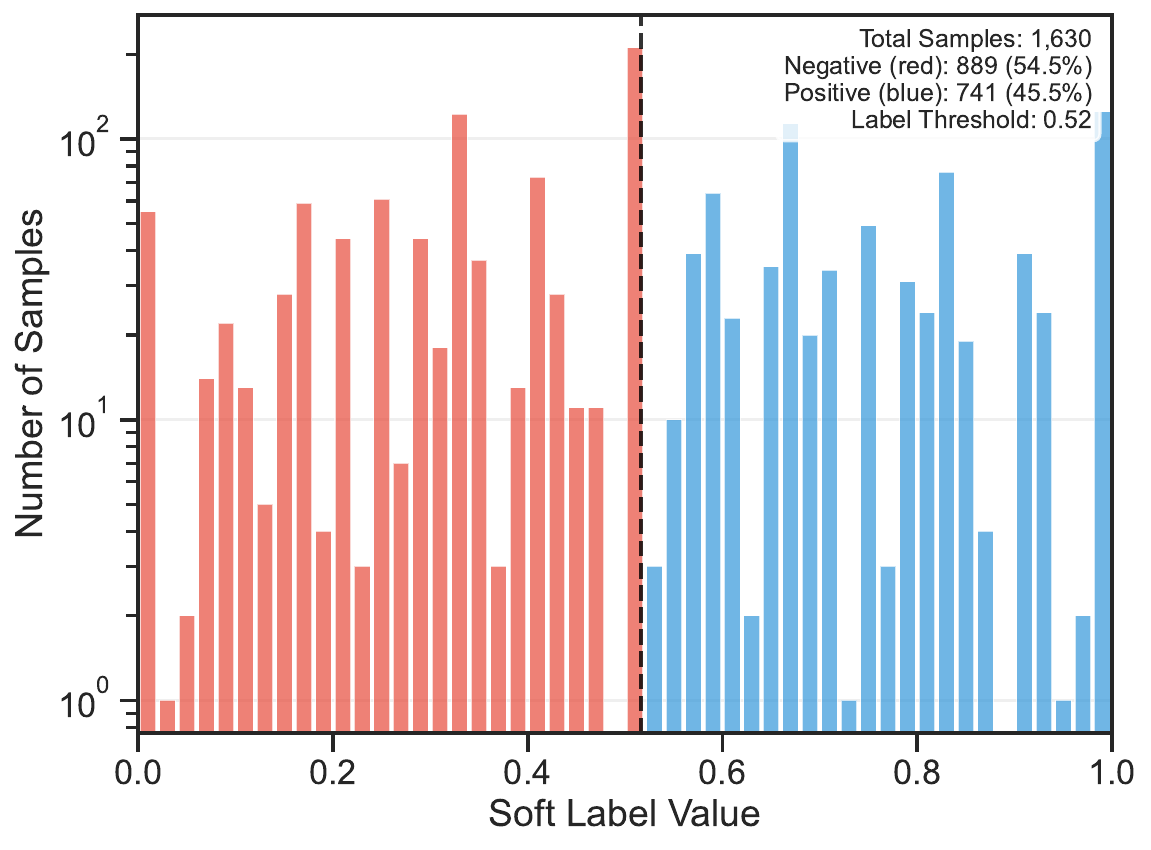}}%
    \subfigure[ENHANCE-Border]{%
      \includegraphics[width=0.32\linewidth]{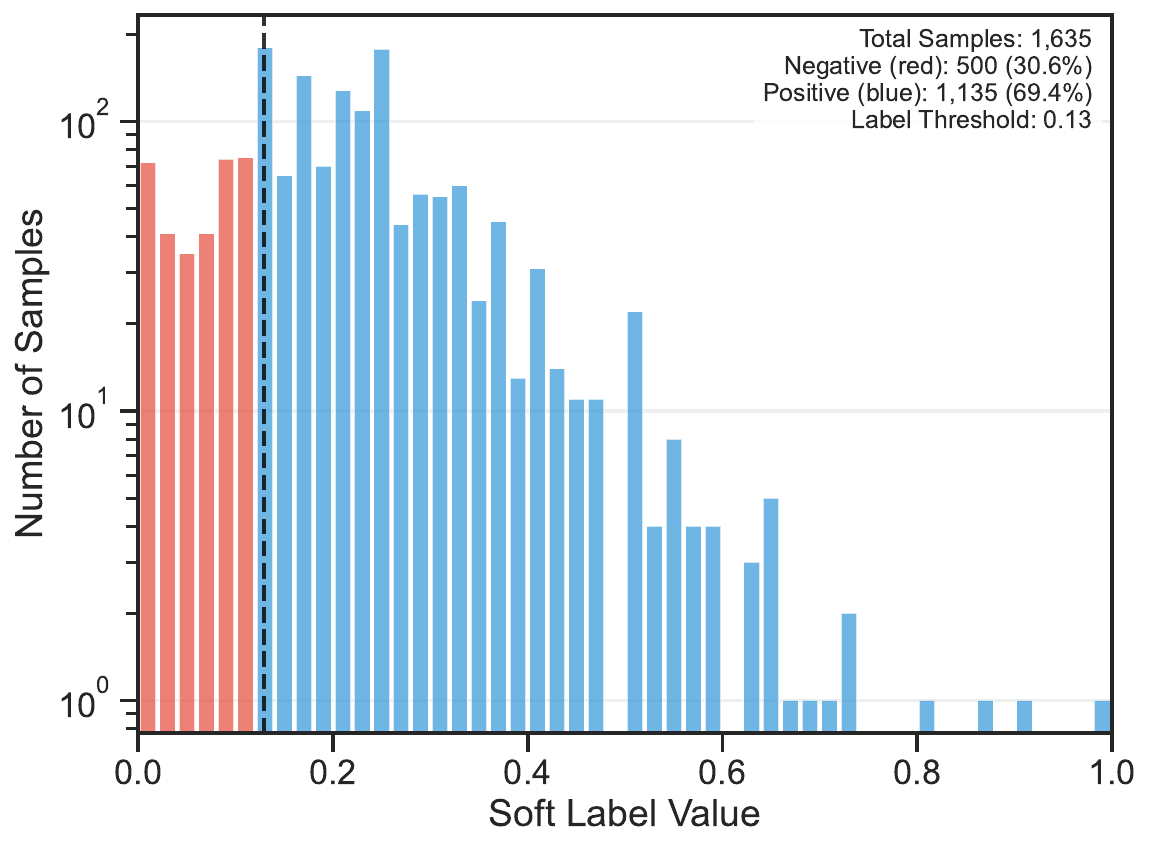}}%
    \subfigure[ENHANCE-Color]{%
      \includegraphics[width=0.32\linewidth]{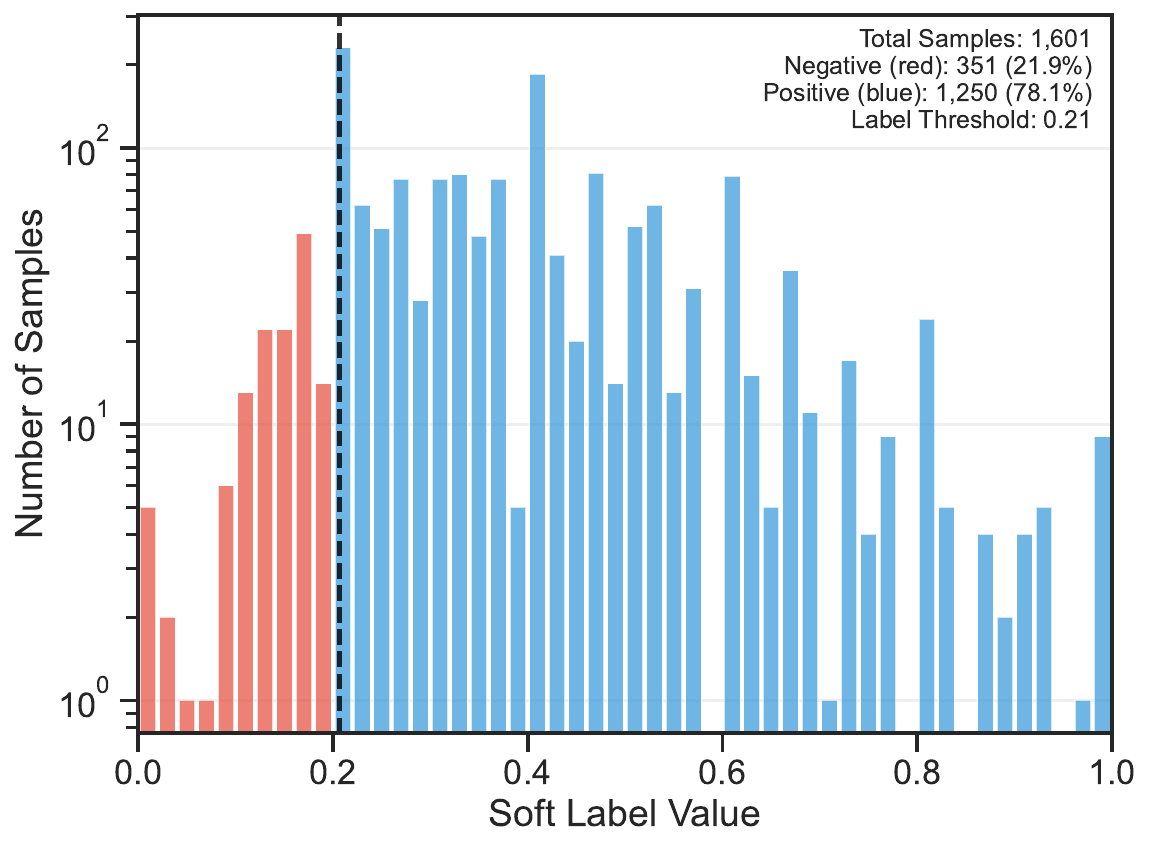}}%
      
    \subfigure[CheXpert-Cardiomegaly]{%
      \includegraphics[width=0.32\linewidth]{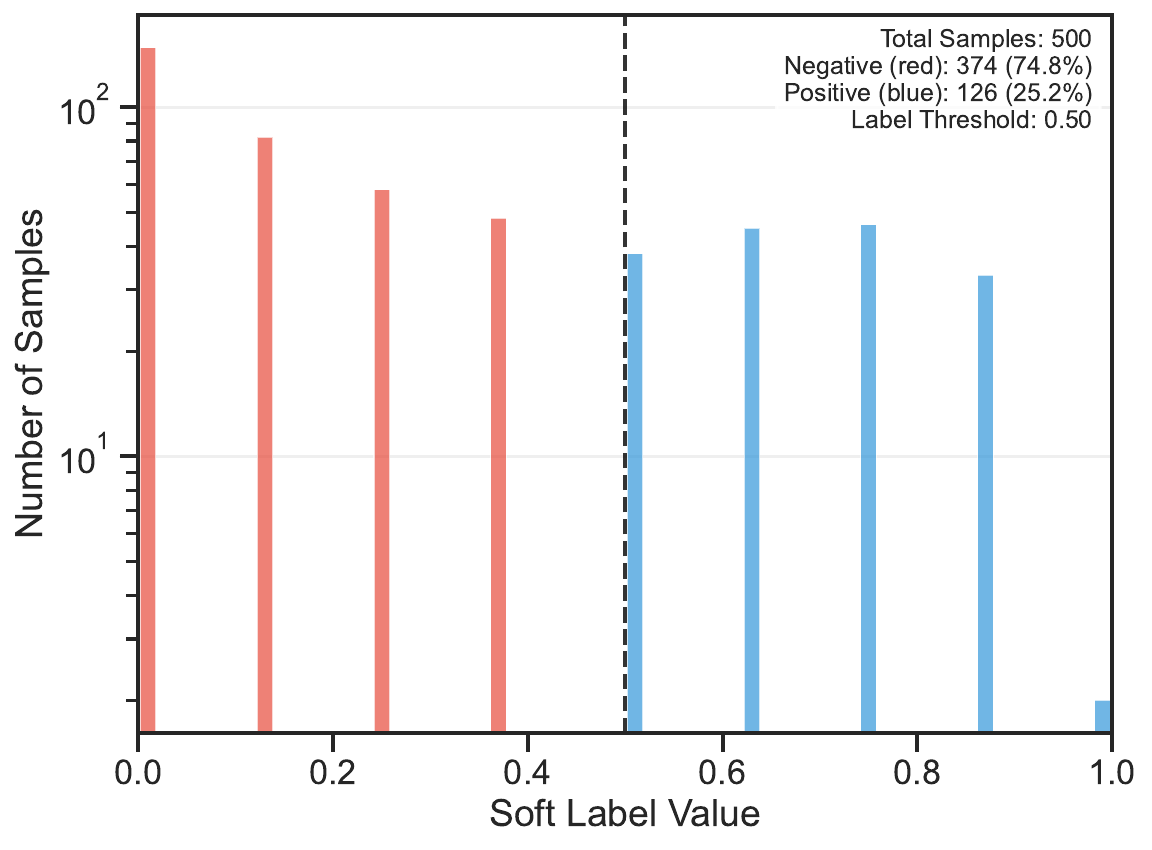}}%
    \subfigure[CheXpert-Pneumothorax]{%
      \includegraphics[width=0.32\linewidth]{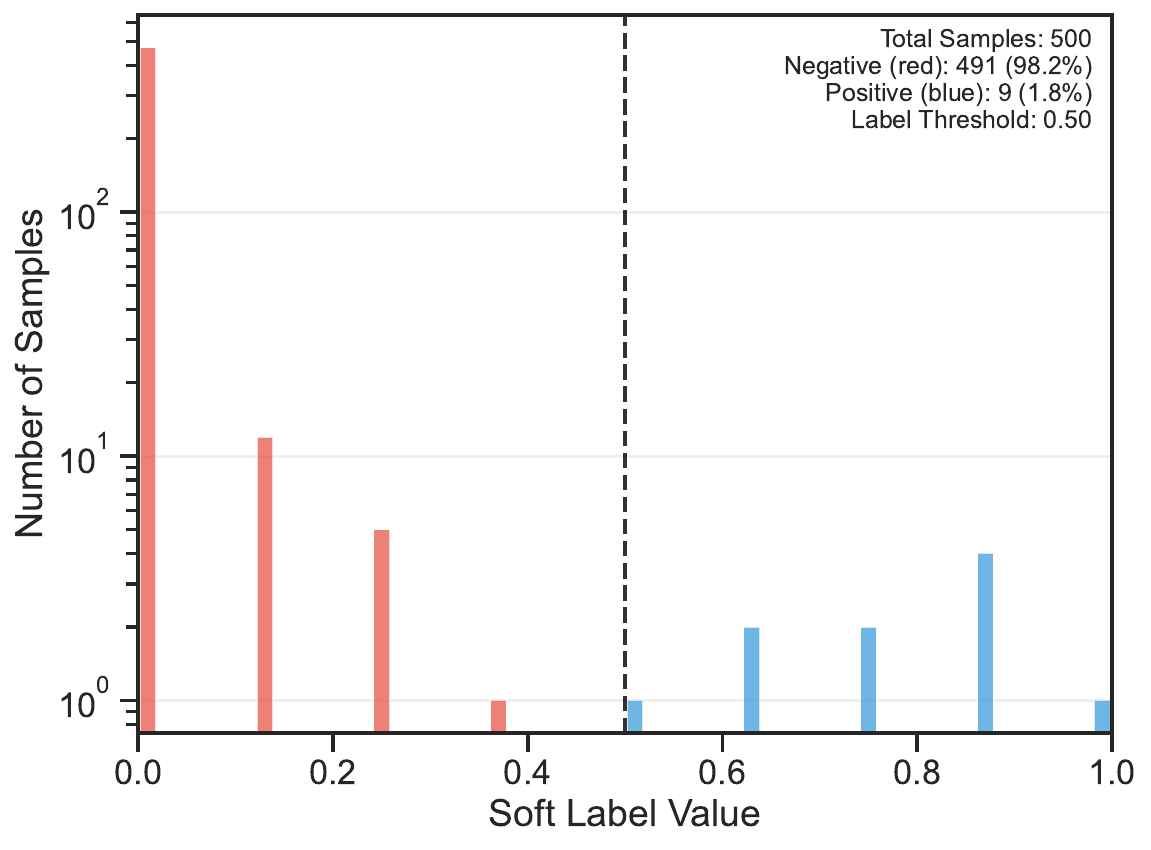}}%
    \subfigure[CheXpert-Pleural Effusion]{%
      \includegraphics[width=0.32\linewidth]{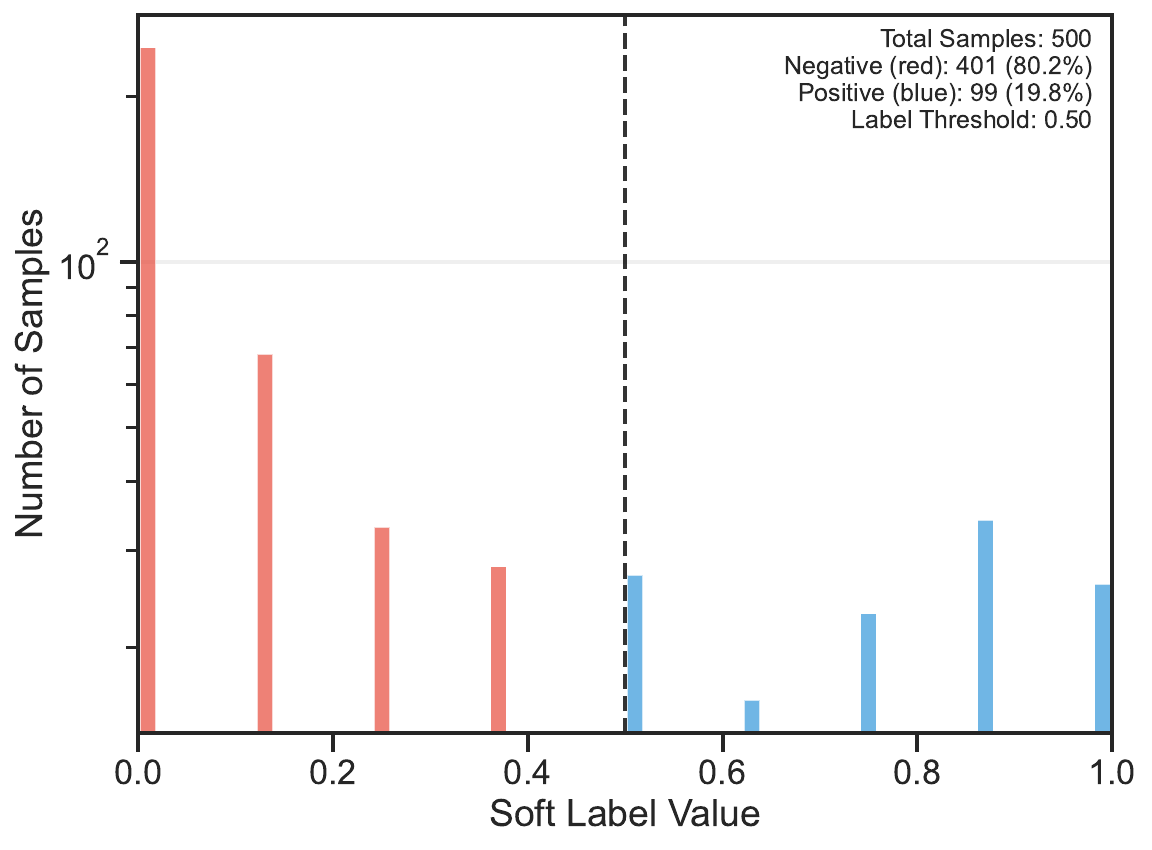}}%
      
    \subfigure[VinDr-Pneumothorax]{%
      \includegraphics[width=0.32\linewidth]{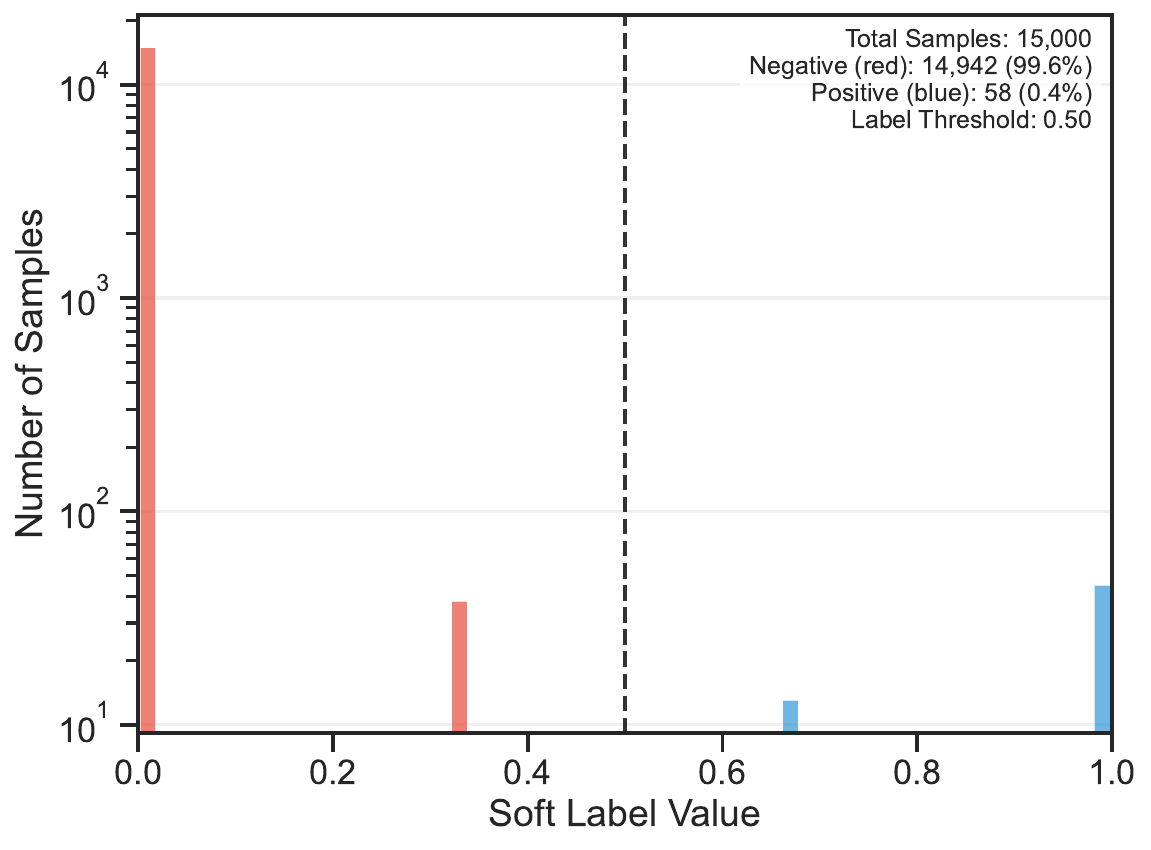}}%
    \subfigure[VinDr-Pulmonary Fibrosis]{%
      \includegraphics[width=0.32\linewidth]{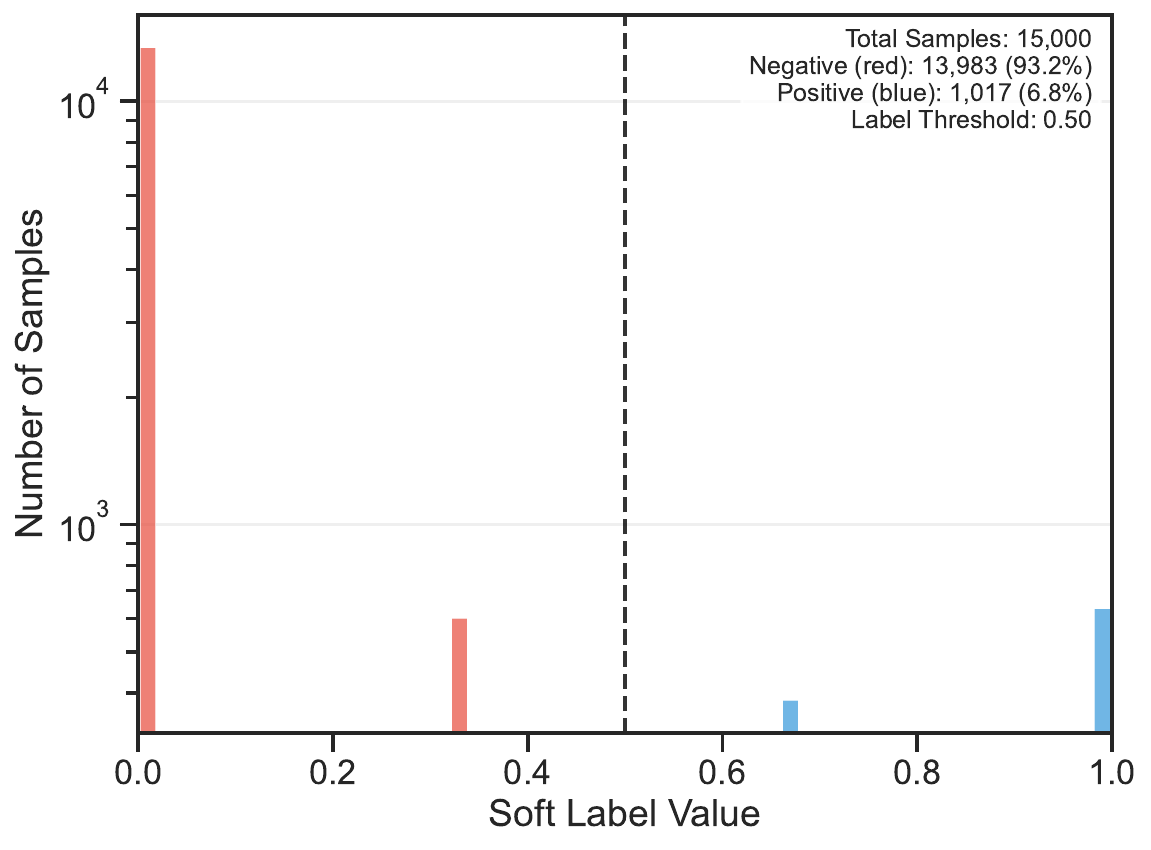}}%
    \subfigure[VinDr-Pleural Thickening]{%
      \includegraphics[width=0.32\linewidth]{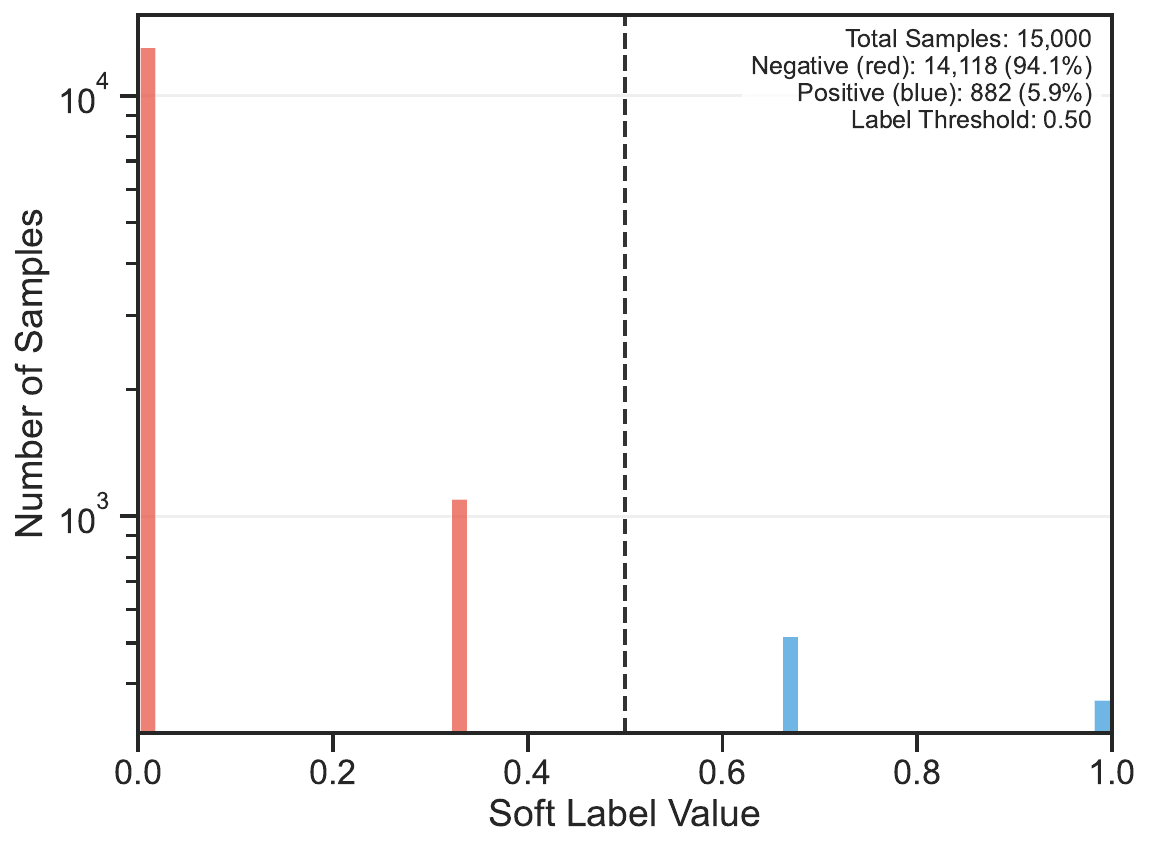}}%
  }
\end{figure*}

\section{Bootstrap analysis details}\label{apd:bootstrap-analysis}
For the bootstrap analysis, we resample the annotations for each image with replacement to generate a new set of labels for the entire dataset. 
We then recalculate all metrics (AUROC, AP, s-AUROC, s-AP) for each model, re-rank the models based on the new scores, and measure the rank correlation with the original order using Spearman's $\rho$ and Kendall's $\tau$. 
We repeat this 1,000 times and use the average rank correlation as an estimate of stability.

\begin{table}[ht]
\centering
\caption{Binomial test p-values comparing the stability (average rank correlation) of ordinary vs. soft metrics over 1,000 bootstrap iterations. Values $<0.05$ indicate the soft metric is significantly more stable. We see this in 8/9 tasks for AP and 3/9 for AUROC.}
\label{tab:bootstrap-pvalues}
\resizebox{\linewidth}{!}{%
\begin{tabular}{lcc}
\toprule
\textbf{Dataset} & \textbf{AP vs. s-AP} & \textbf{AUROC vs. s-AUROC} \\
\midrule
CheXpert-Pleural Effusion & $\phantom{>}$0.0119 & $\phantom{>}$0.2135 \\
CheXpert-Pneumothorax & $>$0.9999 & $>$0.9999 \\
VinDr-Pleural Thickening & $<$0.0001 & $\phantom{>}$0.8019 \\
CheXpert-Cardiomegaly & $<$0.0001 & $\phantom{>}$0.2141 \\
VinDr-Pulmonary Fibrosis & $<$0.0001 & $<$0.0001 \\
VinBigData-Pneumothorax & $\phantom{>}$0.0001 & $\phantom{>}$0.0138 \\
ENHANCE-Color & $\phantom{>}$0.0036 & $\phantom{>}$0.2971 \\
ENHANCE-Border & $<$0.0001 & $\phantom{>}$0.0049 \\
ENHANCE-Asymmetry & $<$0.0001 & $\phantom{>}$0.0625 \\
\bottomrule
\end{tabular}
}
\end{table}

\section{Extended results}\label{apd:ext-results}

\looseness=-1
We provide additional results on the evaluations performed in the main paper. 

\begin{figure*}
    \centering
    \includegraphics[width=\linewidth]{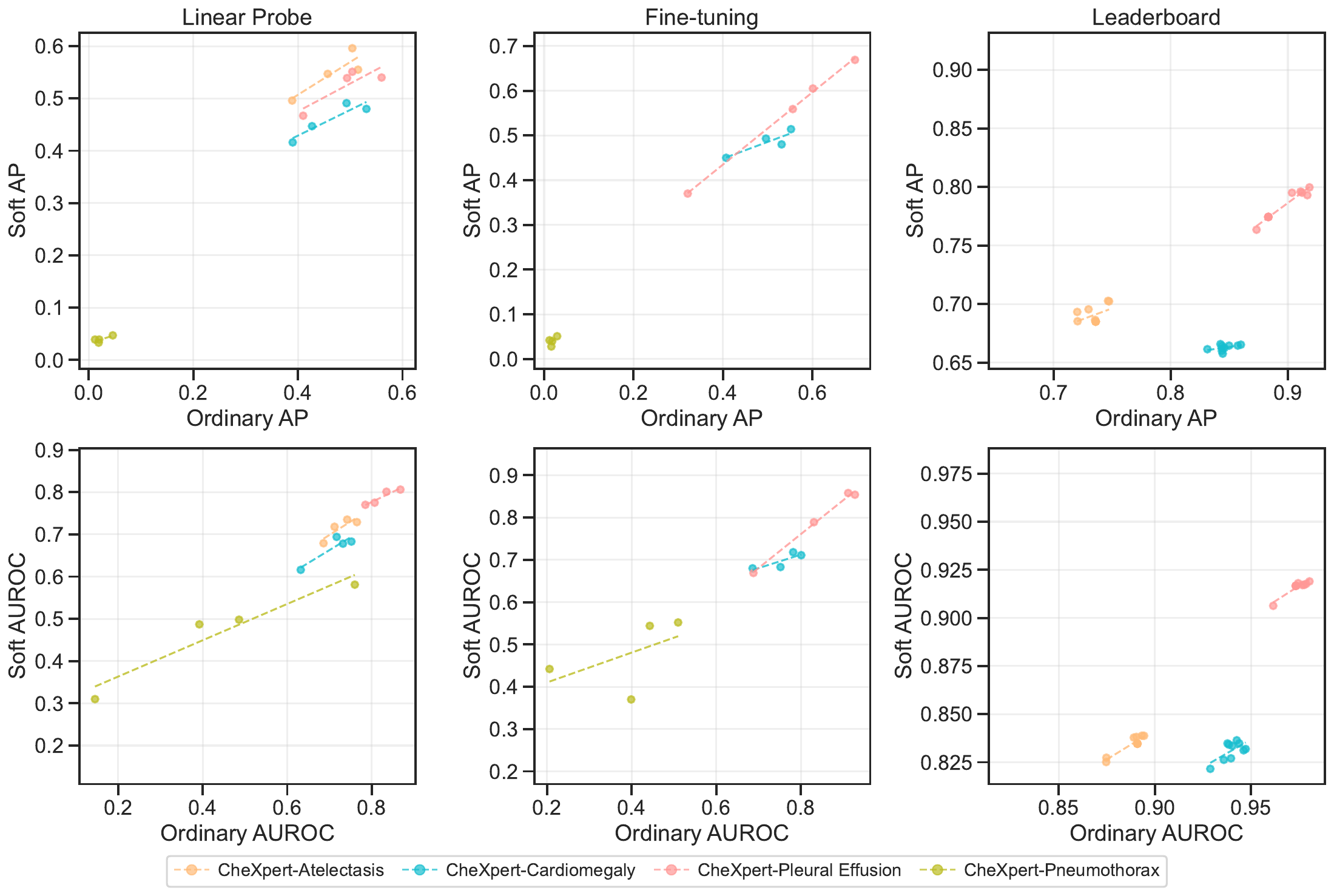}
    \caption{
        Comparison of ordinary and soft metrics on CheXpert for different tasks.
        Evaluation is done for linear probing, fine-tuning, and leaderboard models.
        Dots represent different backbones, and dashed lines indicate Pearson score correlations.
    }
    \label{fig:chexpert-comparison}
\end{figure*}

Table~\ref{tab:OrdinaryVSSoftMetrics} presents the performance evaluation in terms of ordinary and soft metrics for various datasets, specifically for the three attribute detection tasks on \textit{ENHANCE}, five disease prediction tasks on \textit{VinDr}, five disease prediction tasks on \textit{CheXpert}, and the three data quality issue detection tasks on \textit{CleanPatrick}.
To ensure these findings generalize, Table~\ref{tab:finetune-results} shows the results of end-to-end fine-tuning, and Table~\ref{tab:chexternal-reranking} details the re-ranking of external leaderboard models from CheXternal. 
Figure~\ref{fig:chexpert-comparison} visualizes the evaluation of soft versus ordinary metrics for linear probing, fine-tuning, and leaderboard models.

Figure~\ref{fig:bootstrap} shows the results of the bootstrap analysis broken down by dataset, which expands on the aggregated Figure~\ref{fig:stability-avg}.

\begin{table*}[p]
    \centering
    \caption{Comparison of performance measured in terms of ordinary vs.\ soft metrics for different methods and on various datasets.}
    \label{tab:OrdinaryVSSoftMetrics}
    \begin{minipage}[t]{0.49\textwidth}
        \vspace{0pt}
        \centering
        \resizebox{\linewidth}{!}{%
        \begin{tabular}{l cc cc}
        \toprule
            & \multicolumn{2}{c}{\textbf{Ordinary Metrics}}
            & \multicolumn{2}{c}{\textbf{Soft Metrics}} \\

        \cmidrule(lr){2-3}
        \cmidrule(lr){4-5}

        \textbf{Method}
        & AUROC & AP & s-AUROC & s-AP \\

        \midrule
        \multicolumn{5}{l}{\textit{ENHANCE-Asymmetry}} \\
        EfficientNet-b0 & 0.636 & 0.571 & 0.585 & 0.592 \\
        ResNet-50 & 0.656 & 0.621 & 0.593 & 0.618 \\
        VGG-16 & 0.707 & 0.655 & 0.617 & 0.625 \\
        ViT-base & 0.635 & 0.600 & 0.594 & 0.621 \\

        \midrule
        \multicolumn{5}{l}{\textit{ENHANCE-Border}} \\
        EfficientNet-b0 & 0.664 & 0.794 & 0.569 & 0.263 \\
        ResNet-50 & 0.617 & 0.758 & 0.556 & 0.253 \\
        VGG-16 & 0.679 & 0.785 & 0.579 & 0.268 \\
        ViT-base & 0.631 & 0.769 & 0.561 & 0.253 \\

        \midrule
        \multicolumn{5}{l}{\textit{ENHANCE-Color}} \\
        EfficientNet-b0 & 0.709 & 0.890 & 0.558 & 0.431 \\
        ResNet-50 & 0.691 & 0.878 & 0.551 & 0.429 \\
        VGG-16 & 0.694 & 0.886 & 0.560 & 0.443 \\
        ViT-base & 0.656 & 0.849 & 0.533 & 0.408 \\

        \midrule
        \multicolumn{5}{l}{\textit{VinDr-Cardiomegaly}} \\
        EfficientNet-b0 & 0.935 & 0.672 & 0.931 & 0.653 \\
        ResNet-50 & 0.942 & 0.711 & 0.938 & 0.692 \\
        VGG-16 & 0.951 & 0.740 & 0.947 & 0.716 \\
        ViT-base & 0.924 & 0.654 & 0.922 & 0.637 \\

        \midrule
        \multicolumn{5}{l}{\textit{VinDr-Pneumothorax}} \\
        EfficientNet-b0 & 0.958 & 0.216 & 0.903 & 0.191 \\
        ResNet-50 & 0.877 & 0.090 & 0.866 & 0.082 \\
        VGG-16 & 0.928 & 0.232 & 0.924 & 0.224 \\
        ViT-base & 0.915 & 0.093 & 0.912 & 0.096 \\

        \midrule
        \multicolumn{5}{l}{\textit{VinDr-Pleural Effusion}} \\
        EfficientNet-b0 & 0.896 & 0.477 & 0.884 & 0.452 \\
        ResNet-50 & 0.921 & 0.494 & 0.905 & 0.468 \\
        VGG-16 & 0.926 & 0.564 & 0.910 & 0.529 \\
        ViT-base & 0.914 & 0.495 & 0.894 & 0.457 \\

        \midrule
        \multicolumn{5}{l}{\textit{VinDr-Pulmonary Fibrosis}} \\
        EfficientNet-b0 & 0.841 & 0.292 & 0.836 & 0.300 \\
        ResNet-50 & 0.853 & 0.272 & 0.848 & 0.280 \\
        VGG-16 & 0.862 & 0.349 & 0.854 & 0.336 \\
        ViT-base & 0.861 & 0.296 & 0.851 & 0.310 \\

        \midrule
        \multicolumn{5}{l}{\textit{VinDr-Pleural Thickening}} \\
        EfficientNet-b0 & 0.864 & 0.304 & 0.839 & 0.300 \\
        ResNet-50 & 0.870 & 0.302 & 0.849 & 0.297 \\
        VGG-16 & 0.886 & 0.354 & 0.863 & 0.341 \\
        ViT-base & 0.882 & 0.329 & 0.848 & 0.313 \\

        \bottomrule
        \end{tabular}%
        }
    \end{minipage}
    \hfill
    \begin{minipage}[t]{0.49\textwidth}
        \vspace{0pt}
        \centering
        \resizebox{\linewidth}{!}{%
        \begin{tabular}{l cc cc}
        \toprule
            & \multicolumn{2}{c}{\textbf{Ordinary Metrics}}
            & \multicolumn{2}{c}{\textbf{Soft Metrics}} \\

        \cmidrule(lr){2-3}
        \cmidrule(lr){4-5}

        \textbf{Method}
        & AUROC & AP & s-AUROC & s-AP \\

        \midrule
        \multicolumn{5}{l}{\textit{CheXpert-Cardiomegaly}} \\
        EfficientNet-b0 & 0.752 & 0.531 & 0.683 & 0.480 \\
        ResNet-50 & 0.732 & 0.427 & 0.678 & 0.447 \\
        VGG-16 & 0.717 & 0.493 & 0.694 & 0.491 \\
        ViT-base & 0.632 & 0.390 & 0.616 & 0.416 \\

        \midrule
        \multicolumn{5}{l}{\textit{CheXpert-Pneumothorax}} \\
        EfficientNet-b0 & 0.145 & 0.012 & 0.310 & 0.039 \\
        ResNet-50 & 0.392 & 0.020 & 0.487 & 0.039 \\
        VGG-16 & 0.486 & 0.019 & 0.498 & 0.033 \\
        ViT-base & 0.760 & 0.046 & 0.581 & 0.047 \\

        \midrule
        \multicolumn{5}{l}{\textit{CheXpert-Pleural Effusion}} \\
        EfficientNet-b0 & 0.835 & 0.504 & 0.801 & 0.551 \\
        ResNet-50 & 0.868 & 0.560 & 0.806 & 0.540 \\
        VGG-16 & 0.807 & 0.494 & 0.775 & 0.539 \\
        ViT-base & 0.785 & 0.410 & 0.770 & 0.467 \\

        \midrule
        \multicolumn{5}{l}{\textit{CheXpert-Lung Opacity}} \\
        EfficientNet-b0 & 0.847 & 0.828 & 0.813 & 0.813 \\
        ResNet-50 & 0.832 & 0.794 & 0.798 & 0.782 \\
        VGG-16 & 0.842 & 0.813 & 0.804 & 0.804 \\
        ViT-base & 0.820 & 0.799 & 0.799 & 0.796 \\

        \midrule
        \multicolumn{5}{l}{\textit{CheXpert-Atelectasis}} \\
        EfficientNet-b0 & 0.712 & 0.457 & 0.718 & 0.547 \\
        ResNet-50 & 0.765 & 0.515 & 0.729 & 0.555 \\
        VGG-16 & 0.742 & 0.504 & 0.735 & 0.596 \\
        ViT-base & 0.686 & 0.389 & 0.679 & 0.496 \\

        \midrule
        \multicolumn{5}{l}{\textit{CleanPatrick: Off-topic Samples}} \\
        SelfClean & 0.669 & 0.145 & 0.676 & 0.123 \\
        ECOD & 0.757 & 0.162 & 0.754 & 0.134 \\
        IForest & 0.773 & 0.159 & 0.774 & 0.129 \\
        DeepSVD & 0.728 & 0.130 & 0.714 & 0.079 \\

        \midrule
        \multicolumn{5}{l}{\textit{CleanPatrick: Near Duplicates}} \\
        SelfClean & 0.917 & 0.879 & 0.888 & 0.111 \\
        pHash & 0.505 & 0.316 & 0.493 & 0.025 \\
        SSIM & 0.491 & 0.305 & 0.495 & 0.025 \\

        \midrule
        \multicolumn{5}{l}{\textit{CleanPatrick: Label Errors}} \\
        SelfClean & 0.572 & 0.265 & 0.804 & 0.018 \\
        CLearning & 0.477 & 0.216 & 0.494 & 0.002 \\
        FINE & 0.469 & 0.205 & 0.396 & 0.001 \\
        BHN & 0.547 & 0.239 & 0.545 & 0.002 \\

        \bottomrule
        \end{tabular}%
        }
    \end{minipage}
\end{table*}

\begin{table}[ht]
\centering
\caption{Performance comparison of end-to-end fine-tuned models on CheXpert tasks. Rank-flipping is observed (e.g., VGG-16 vs. ResNet-50 on Cardiomegaly AP/s-AP).}
\label{tab:finetune-results}
\resizebox{\linewidth}{!}{%
\begin{tabular}{lcccc}
\toprule
\textbf{Model} & \textbf{AUROC} & \textbf{s-AUROC} & \textbf{AP} & \textbf{s-AP} \\
\midrule
\textit{Cardiomegaly} \\
EfficientNet-b0 & 0.752 & 0.683 & 0.531 & 0.480 \\
ResNet-50 & 0.801 & 0.711 & 0.552 & 0.514 \\
VGG-16 & 0.782 & 0.718 & 0.496 & 0.493 \\
ViT-base & 0.686 & 0.680 & 0.407 & 0.450 \\
\midrule
\textit{Pneumothorax} \\
EfficientNet-b0 & 0.206 & 0.442 & 0.012 & 0.042 \\
ResNet-50 & 0.443 & 0.544 & 0.018 & 0.040 \\
VGG-16 & 0.510 & 0.552 & 0.029 & 0.051 \\
ViT-base & 0.399 & 0.370 & 0.016 & 0.028 \\
\midrule
\textit{Pleural Effusion} \\
EfficientNet-b0 & 0.831 & 0.789 & 0.556 & 0.559 \\
ResNet-50 & 0.928 & 0.854 & 0.695 & 0.669 \\
VGG-16 & 0.912 & 0.858 & 0.601 & 0.605 \\
ViT-base & 0.688 & 0.669 & 0.321 & 0.370 \\
\bottomrule
\end{tabular}
}
\end{table}

\begin{table*}[ht]
\centering
\caption{Re-ranking of CheXternal leaderboard models when switching from ordinary to soft metrics.}
\label{tab:chexternal-reranking}
\resizebox{0.8\textwidth}{!}{%
\begin{tabular}{lcccccccc}
\toprule
& \multicolumn{2}{c}{\textbf{AUROC}} 
& \multicolumn{2}{c}{\textbf{s-AUROC}} 
& \multicolumn{2}{c}{\textbf{AP}} 
& \multicolumn{2}{c}{\textbf{s-AP}} \\
\cmidrule(lr){2-3} \cmidrule(lr){4-5} \cmidrule(lr){6-7} \cmidrule(lr){8-9}
\textbf{Model} 
    & Score & O. Rank 
    & Score & O. Rank  
    & Score & O. Rank  
    & Score & O. Rank \\
\midrule
\textit{Atelectasis} \\
jfaboy & 0.894 & 1 & 0.839 & 2 & 0.747 & 1 & 0.702 & 2 \\
ngango3 & 0.893 & 2 & 0.839 & 1 & 0.747 & 2 & 0.703 & 1 \\
uestc & 0.891 & 3 & 0.835 & 5 & 0.736 & 3 & 0.685 & 7 \\
drnet & 0.891 & 3 & 0.835 & 5 & 0.736 & 3 & 0.685 & 7 \\
sensexdr & 0.891 & 3 & 0.835 & 5 & 0.736 & 3 & 0.685 & 7 \\
ihil & 0.891 & 3 & 0.835 & 5 & 0.736 & 3 & 0.685 & 7 \\
ngango2 & 0.890 & 7 & 0.838 & 3 & 0.730 & 8 & 0.696 & 3 \\
hieupham & 0.889 & 8 & 0.838 & 4 & 0.720 & 10 & 0.693 & 4 \\
desmond & 0.875 & 9 & 0.827 & 9 & 0.720 & 9 & 0.685 & 6 \\
yww211 & 0.875 & 10 & 0.825 & 10 & 0.735 & 7 & 0.687 & 5 \\
\midrule
\textit{Cardiomegaly} \\
sensexdr & 0.947 & 1 & 0.832 & 6 & 0.599 & 10 & 0.665 & 2 \\
ihil & 0.946 & 2 & 0.831 & 7 & 0.857 & 1 & 0.665 & 3 \\
ngango3 & 0.944 & 3 & 0.835 & 2 & 0.844 & 7 & 0.664 & 5 \\
hieupham & 0.943 & 4 & 0.836 & 1 & 0.851 & 2 & 0.666 & 1 \\
desmond & 0.940 & 5 & 0.834 & 5 & 0.844 & 6 & 0.662 & 7 \\
drnet & 0.940 & 6 & 0.827 & 8 & 0.844 & 8 & 0.660 & 9 \\
ngango2 & 0.939 & 7 & 0.834 & 4 & 0.840 & 9 & 0.662 & 8 \\
yww211 & 0.938 & 8 & 0.835 & 3 & 0.850 & 3 & 0.665 & 3 \\
uestc & 0.936 & 9 & 0.826 & 9 & 0.846 & 4 & 0.663 & 6 \\
jfaboy & 0.929 & 10 & 0.822 & 10 & 0.844 & 5 & 0.658 & 10 \\
\midrule
\textit{Consolidation} \\
jfaboy & 0.927 & 1 & 0.853 & 1 & 0.451 & 1 & 0.402 & 1 \\
uestc & 0.921 & 2 & 0.848 & 3 & 0.408 & 2 & 0.397 & 2 \\
drnet & 0.921 & 2 & 0.848 & 3 & 0.408 & 2 & 0.397 & 2 \\
sensexdr & 0.921 & 2 & 0.848 & 3 & 0.408 & 2 & 0.397 & 2 \\
ihil & 0.921 & 2 & 0.848 & 3 & 0.408 & 2 & 0.397 & 2 \\
yww211 & 0.918 & 6 & 0.851 & 2 & 0.354 & 7 & 0.392 & 6 \\
ngango3 & 0.912 & 7 & 0.846 & 8 & 0.382 & 6 & 0.384 & 7 \\
desmond & 0.905 & 8 & 0.847 & 7 & 0.330 & 8 & 0.383 & 8 \\
ngango2 & 0.892 & 9 & 0.835 & 9 & 0.279 & 10 & 0.353 & 10 \\
hieupham & 0.891 & 10 & 0.831 & 10 & 0.304 & 9 & 0.357 & 9 \\
\midrule
\textit{Edema} \\
drnet & 0.935 & 1 & 0.871 & 1 & 0.706 & 1 & 0.633 & 1 \\
sensexdr & 0.935 & 1 & 0.871 & 1 & 0.706 & 1 & 0.633 & 1 \\
ihil & 0.935 & 1 & 0.871 & 1 & 0.706 & 1 & 0.633 & 1 \\
uestc & 0.935 & 4 & 0.869 & 4 & 0.694 & 4 & 0.628 & 4 \\
yww211 & 0.932 & 5 & 0.865 & 5 & 0.633 & 8 & 0.598 & 7 \\
desmond & 0.929 & 6 & 0.859 & 8 & 0.637 & 6 & 0.597 & 8 \\
ngango2 & 0.926 & 7 & 0.864 & 6 & 0.646 & 5 & 0.606 & 5 \\
hieupham & 0.924 & 8 & 0.862 & 7 & 0.637 & 7 & 0.603 & 6 \\
jfaboy & 0.922 & 9 & 0.856 & 10 & 0.566 & 10 & 0.566 & 10 \\
ngango3 & 0.911 & 10 & 0.857 & 9 & 0.590 & 9 & 0.588 & 9 \\
\midrule
\textit{Pleural Effusion} \\
hieupham & 0.979 & 1 & 0.920 & 1 & 0.908 & 4 & 0.796 & 1 \\
ngango3 & 0.979 & 2 & 0.918 & 3 & 0.911 & 3 & 0.796 & 2 \\
jfaboy & 0.978 & 3 & 0.917 & 4 & 0.917 & 1 & 0.793 & 5 \\
ngango2 & 0.977 & 4 & 0.917 & 5 & 0.350 & 10 & 0.795 & 3 \\
yww211 & 0.974 & 5 & 0.918 & 2 & 0.912 & 2 & 0.795 & 3 \\
uestc & 0.973 & 6 & 0.917 & 6 & 0.883 & 5 & 0.774 & 6 \\
drnet & 0.973 & 6 & 0.917 & 6 & 0.883 & 5 & 0.774 & 6 \\
sensexdr & 0.973 & 6 & 0.917 & 6 & 0.883 & 5 & 0.774 & 6 \\
ihil & 0.973 & 6 & 0.917 & 6 & 0.883 & 5 & 0.774 & 6 \\
desmond & 0.962 & 10 & 0.906 & 10 & 0.849 & 9 & 0.764 & 10 \\
\bottomrule
\end{tabular}
} 
\end{table*}

\begin{figure*}[htbp]
\floatconts
  {fig:bootstrap}
  {\caption{
    Correlation coefficient of model rankings produced by both ordinary and soft metrics averaged over 1,000 bootstrapped samples, which represent random annotation draws.
  }}
  {%
    \subfigure[Kendall's $\tau$]{%
      \includegraphics[width=0.5\linewidth]{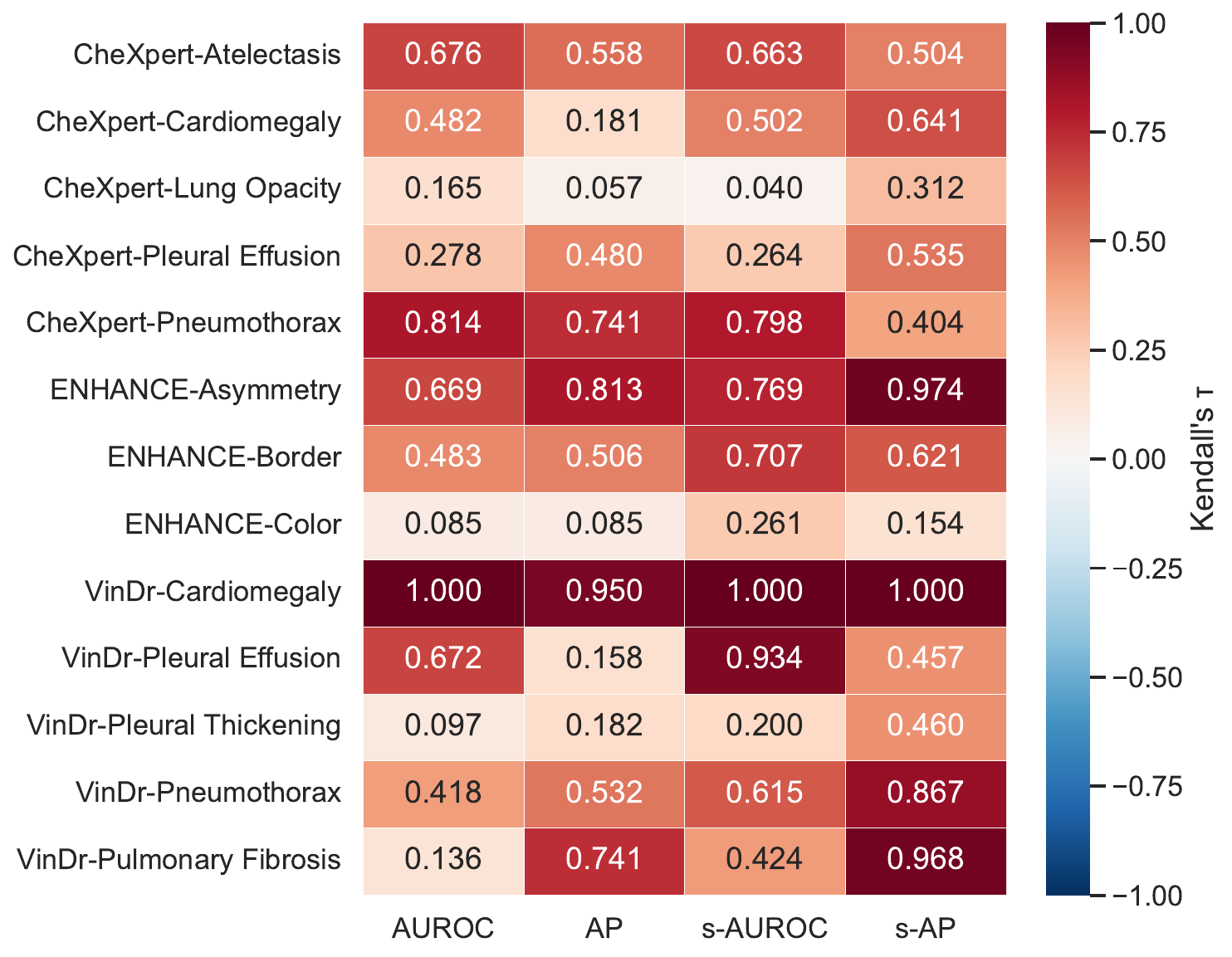}}%
    \subfigure[Spearman's $\rho$]{%
      \includegraphics[width=0.5\linewidth]{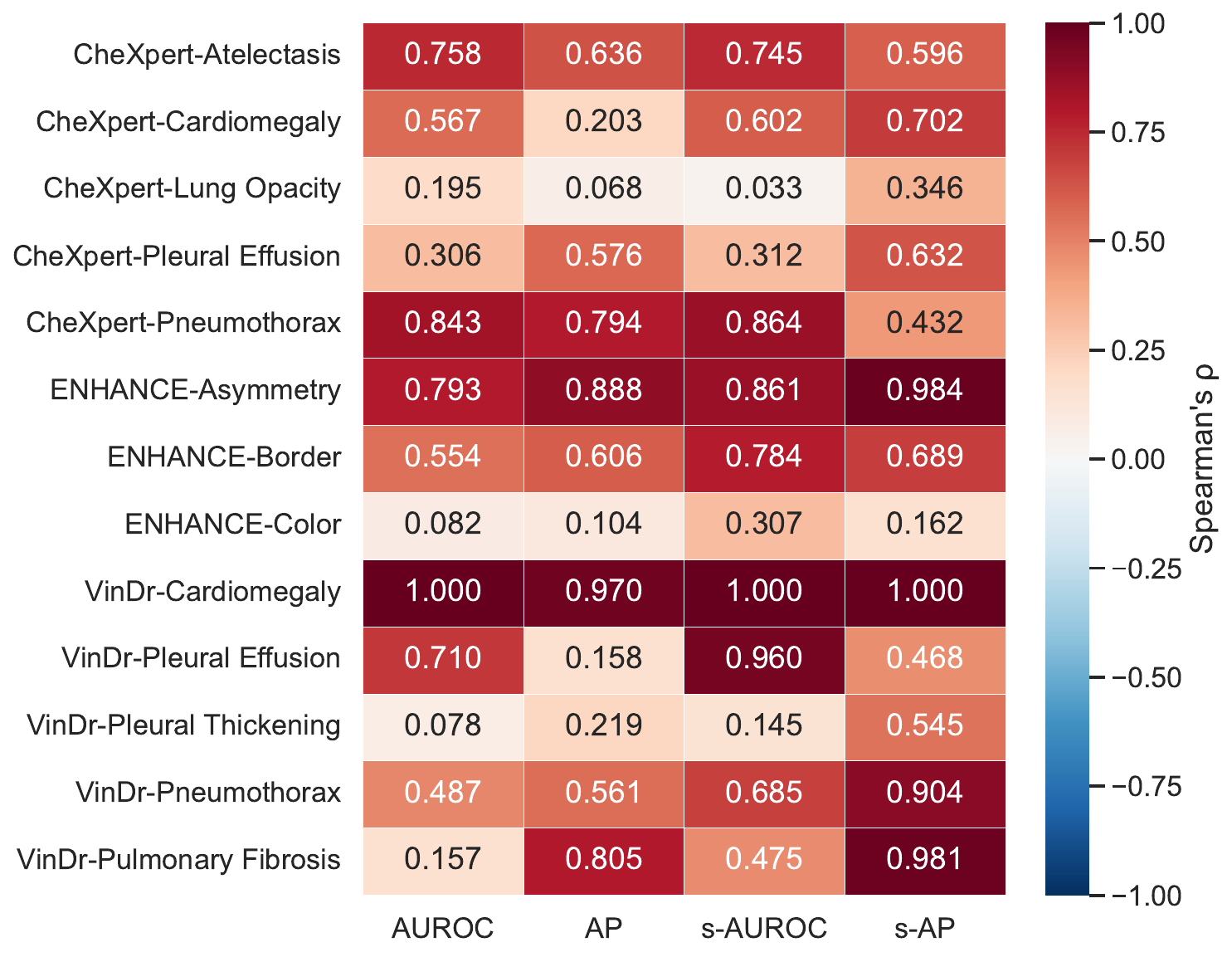}}%
  }
\end{figure*}

\section{Description of cleaning approaches}\label{apd:data-quality-methods}

We briefly outline the methods we evaluated to detect off-topic samples, near duplicates, and label errors. 
For implementation details, please refer to the cited papers.
For hyperparameters, we follow the same strategy as \textit{CleanPatrick} and use all methods as implemented in the benchmark \citep{groger2025cleanpatrick}.

\subsection{Off-topic samples}

\textbf{Isolation Forest (IForest)}: Unsupervised anomaly detection by randomly partitioning features, where points with short average path length are flagged as outliers \citep{liu_isolation_2008}.
\newline
\textbf{ECOD}: Tail-sensitive, distribution-free outlier scoring based on empirical CDFs computed per feature \citep{li_ecod_2022}.
\newline
\textbf{DeepSVD}: One-class deep anomaly detection that learns compact representations and flags samples far from the learned support (deep one-class objective) \citep{ruff2018deep}.

\subsection{Near duplicates}

\textbf{pHash}: Perceptual hashing; small Hamming distance between hashes indicates visually similar images \citep{zauner2010implementation}.
\newline
\textbf{SSIM}: Local comparison of luminance, contrast, and structure, where high mean SSIM signals near-duplicates \citep{wang_image_2004}.

\subsection{Label errors}

\textbf{Confident Learning (CLearning)}: Estimates class-conditional noise from model predictions to identify likely mislabeled examples \citep{northcutt_confident_2022}.
\newline
\textbf{FINE}: Influence-based ranking that prioritizes examples whose labels appear inconsistent with the learned decision boundary \citep{kim2021fine}.
\newline
\textbf{BHN}: Bayesian uncertainty-driven scoring that flags candidates for relabeling based on predictive uncertainty \citep{yu2023delving}.

\subsection{Multiple issue types}

\textbf{SelfClean}: Uses context-aware self-supervised embeddings and neighborhood consistency to rank likely issues \citep{groger_selfclean_2024}.
The proposed method can be used with a human-in-the-loop or thresholded for automation.
Here full automation is used.

\end{document}